\documentclass[pdflatex,sn-mathphys-num]{sn-jnl}

\usepackage{graphicx}%
\usepackage{multirow}%
\usepackage{amsmath,amssymb,amsfonts}%
\usepackage{amsthm}%
\usepackage{mathrsfs}%
\usepackage[title]{appendix}%
\usepackage{xcolor}%
\usepackage{textcomp}%
\usepackage{manyfoot}%
\usepackage{booktabs}%
\usepackage{algorithm}%
\usepackage{algorithmicx}%
\usepackage{algpseudocode}%
\usepackage{listings}%

\theoremstyle{thmstyleone}%
\theoremstyle{thmstyletwo}%

\theoremstyle{thmstylethree}%

\begin{document}

\title[Humanoid Robot-Assisted Sinus Surgery]{Humanoid Robots as First Assistants in Endoscopic Surgery}


\author*[1]{\fnm{Sue Min} \sur{Cho}} \email{scho72@jhu.edu}
\equalcont{These authors contributed equally to this work.}

\author[1]{\fnm{Jan Emily} \sur{Mangulabnan}} \email{jmangul1@jh.edu}
\equalcont{These authors contributed equally and reserve the right to list themselves first on professional documents.}

\author[1]{\fnm{Han} \sur{Zhang}} \email{hzhan206@jhu.edu}
\equalcont{These authors contributed equally and reserve the right to list themselves first on professional documents.}
\author[1]{\fnm{Zhekai} \sur{Mao}} \email{zmao17@jh.edu} 

\author[3]{\fnm{Yufan} \sur{He}} \email{yufanh@nvidia.com}
\author[3]{\fnm{Pengfei} \sur{Guo}} \email{pengfeig@nvidia.com}
\author[3]{\fnm{Daguang} \sur{Xu}} \email{daguangx@nvidia.com}

\author[1]{\fnm{Gregory} \sur{Hager}} \email{hager@jhu.edu}
\author[2]{\fnm{Masaru} \sur{Ishii}} \email{mishii3@jhmi.edu}
\author[1,2]{\fnm{Mathias} \sur{Unberath}} \email{unberath@jhu.edu}

\affil*[1]{\orgname{Johns Hopkins University}, \orgaddress{\city{Baltimore}, \state{MD}, \country{USA}}}

\affil[2]{\orgname{Johns Hopkins Medical Institutions}, \orgaddress{\city{Baltimore}, \state{MD}, \country{USA}}}

\affil[3]{\orgname{NVIDIA Corporation}, \orgaddress{\city{Bethesda}, \state{MD}, \country{USA}}}


\abstract{Humanoid robots have become a focal point of technological ambition, with claims of surgical capability within years in mainstream discourse. These projections are aspirational yet lack empirical grounding. To date, no humanoid has assisted a surgeon through an actual procedure, let alone performed one. The work described here breaks this new ground. Here we report a proof of concept in which a teleoperated Unitree G1 provided endoscopic visualization while an attending otolaryngologist performed a cadaveric sphenoidectomy. The procedure was completed successfully, with stable visualization maintained throughout. Teleoperation allowed assessment of whether the humanoid form factor could meet the physical demands of surgical assistance in terms of sustenance and precision; the cognitive demands were satisfied -- for now -- by the operator. Post-procedure analysis identified engineering targets for clinical translation, alongside near-term opportunities such as autonomous diagnostic scoping. This work establishes form-factor feasibility for humanoid surgical assistance while identifying challenges for continued development.
}




\maketitle

\section{Main}\label{main}
The Operating Room (OR) is a complex, crowded environment where surgeons, nurses, anesthesiologists, and trainees occupy carefully coordinated positions around the patient, surrounded by instruments, imaging systems, and support equipment\cite{kleinbeck_neural_2024,özsoy2025mmorlargemultimodaloperating,zhang2025didjustsee,perez2025orworkflow,zhang2025twinor,StraightTrack}. Within this space, surgical robots have been increasingly assuming specialized roles. Many provide \textit{superhuman capabilities}: tremor filtration, motion scaling, or access to anatomical regions beyond manual reach~\cite{cepolina2022introductory}. Others augment -- rather than exceed -- human capability. Bedside camera holders such as SoloAssist, FreeHand, and Maestro with ScoPilot maintain stable endoscopic visualization without assistant fatigue, reducing workload and freeing skilled personnel~\cite{beasley2012medical,mercoli2026safety}.

Yet current robotic systems, whether superhuman or augmentative, share a common constraint: the operating room must adapt to them. Docking, draping, and positioning follow device-specific protocols; team members reposition to accommodate robotic footprints; workflows bend around fixed integration pathways.

Humanoid robots, in contrast, offer a fundamentally different approach: they embody \textit{human versatility}. Designed to function in environments built for people, humanoids can navigate the same spaces and perform a range of support or preparatory tasks without extensive customization~\cite{tong2024advancements}. Their general-purpose dexterity and anthropomorphic form could allow them to integrate naturally into OR workflows as assistants, equipment managers, or setup facilitators, where adaptability and context awareness often matter more than submillimeter accuracy.

Growing interest in humanoid robots for medical contexts indicates the potential extension towards general-purpose, adaptable automation in healthcare~\cite{ruiz2025care}. The anthropomorphic design and human-compatible dexterity embodied by humanoids suggest opportunities to support a wide range of clinical activities, particularly those requiring situational awareness and physical interaction within human-designed spaces~\cite{ozturkcan2022humanoid}. 

Recent work has begun exploring this possibility. In mannequin experiments, teleoperated humanoid robots have been shown capable of executing various medical tasks, from physical examinations to emergency interventions~\cite{atar2025humanoids}. Demonstrations have included auscultation, bag-valve-mask ventilation, and ultrasound-guided procedures. These demonstrations show that the humanoid form factor can operate in clinical environments and carry out medically relevant actions.

But these tasks differ from surgical assistance in important ways. They tend to be brief, performed on phantoms, and do not involve maneuvering instruments within narrow anatomical spaces in close proximity to delicate anatomy. Assisting in surgery often requires precisely this: sustained, stable positioning of tools within confined bodily cavities. Before investing heavily in autonomous systems, which require substantial work in perception, planning, and control, foundational questions need to be answered. Can the humanoid form factor operate within a confined surgical workspace? Maintain stable instrument positioning? Enable a surgeon to complete a real procedure? And what limitations and development needs emerge from attempting it?

\begin{figure}[t]
    \centering
    \includegraphics[width=0.8\linewidth]{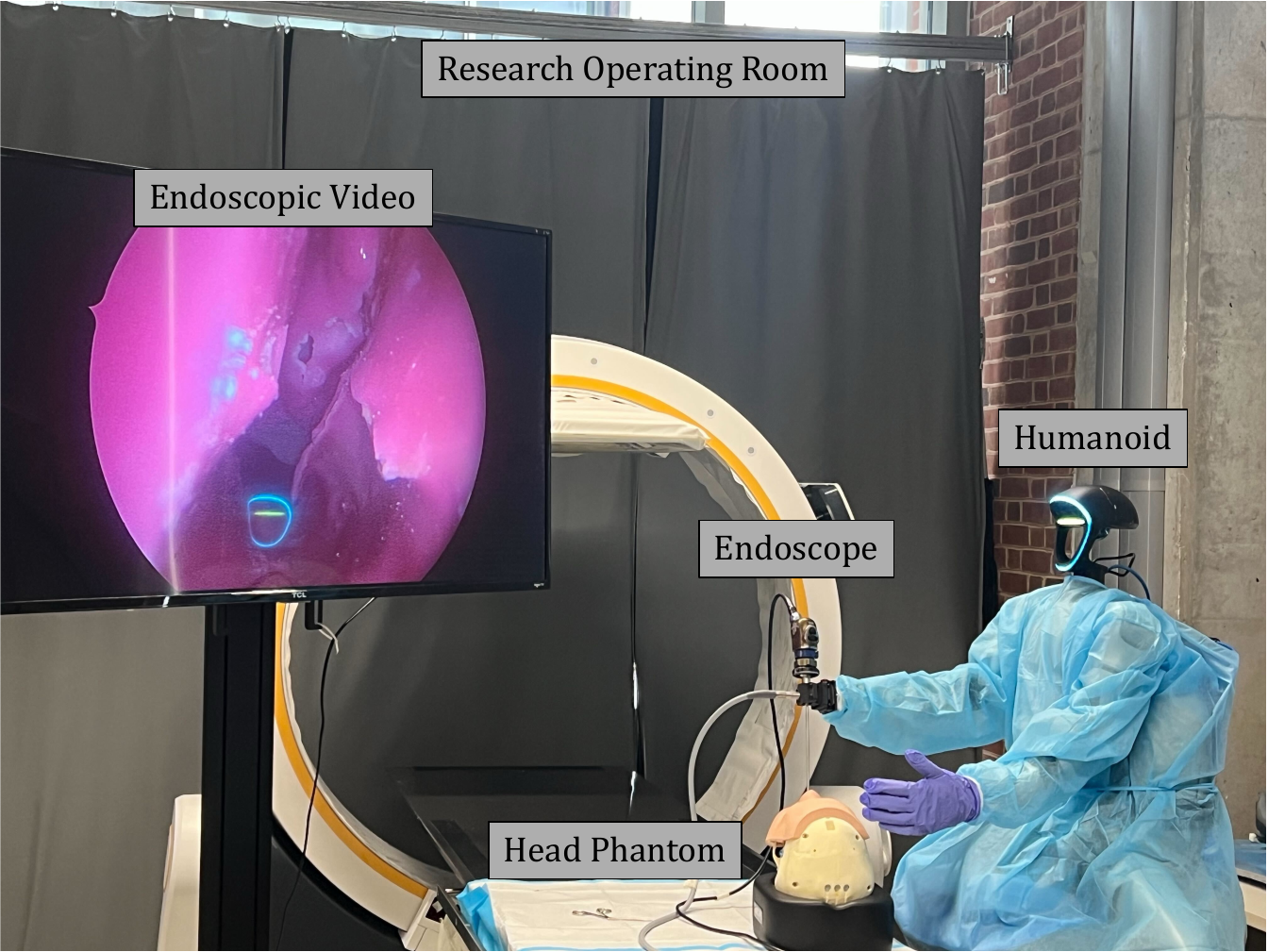}
    \caption{Experimental configuration of the humanoid-assisted endoscopic setup shown using a phantom specimen to illustrate spatial arrangement. The Unitree G1 humanoid robot provided visualization with a rigidly mounted endoscope. The cadaveric sphenoidectomy used for feasibility evaluation is shown in Supplementary Fig.~\ref{fig:cadaver}}
    \label{fig:overview}
\end{figure}

We report on a proof of concept to address these questions. Teleoperation allowed us to assess whether the form factor could meet the physical demands of surgical assistance --sustenance and precision -- while a human operator supplied the cognitive demands of perception, anticipation, and responsive adjustment to surgeon cues. A teleoperated Unitree G1 humanoid robot provided endoscopic visualization during a sphenoidectomy performed by an attending otolaryngologist on a cadaveric specimen (Fig.~\ref{fig:overview}). Following the procedure, we conducted a semi-structured interview to characterize the surgeon's experience across five domains: task performance, form factor and workspace, clinical applications, trust and safety, and interaction and control.

\begin{figure}
    \centering
    \includegraphics[width=1.0\linewidth]{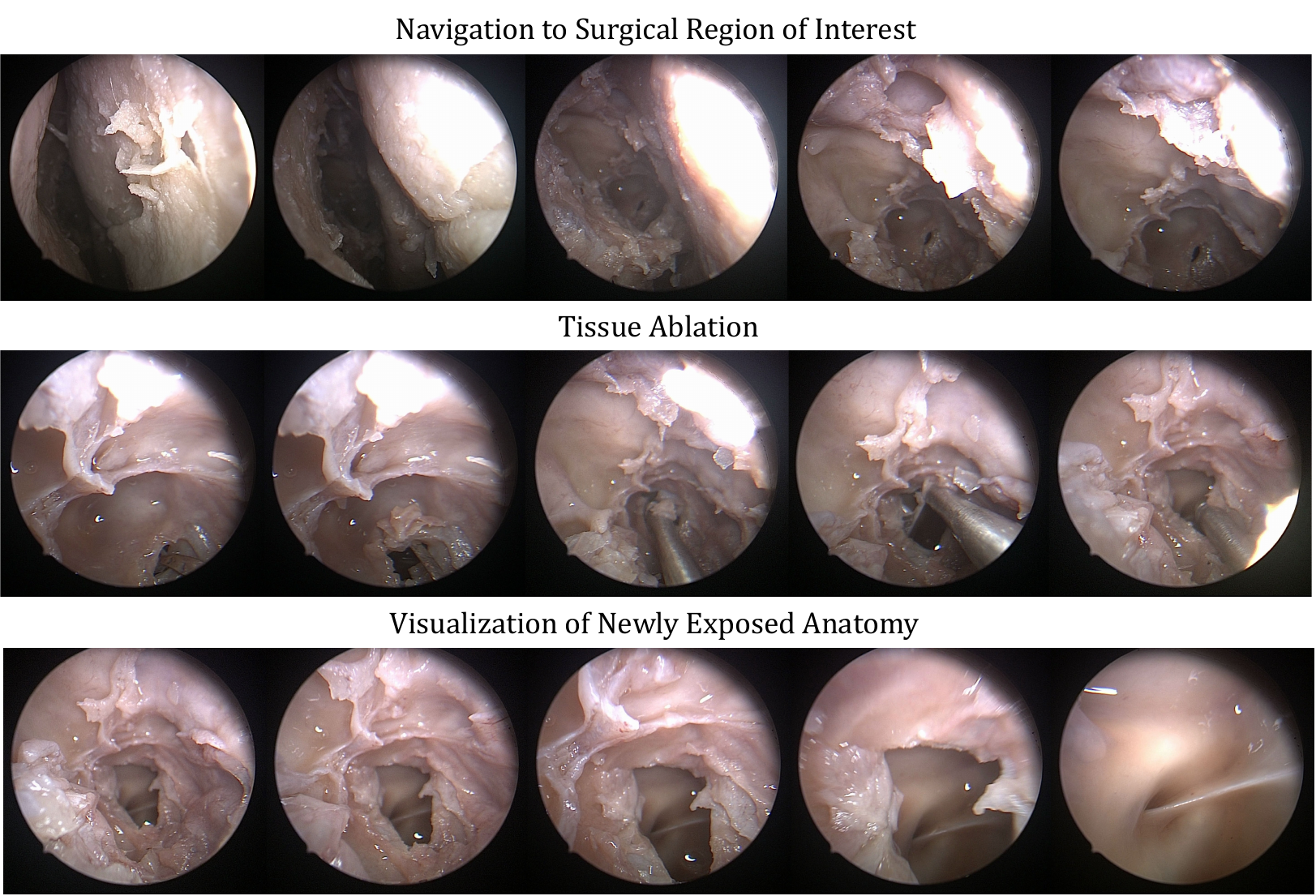}
    \caption{Endoscopic views from the cadaveric sphenoidectomy performed under teleoperated humanoid visualization. Procedural progression is shown from navigation to the surgical region of interest, through tissue ablation by the surgeon, to visualization of newly exposed anatomical structures.}
    \label{fig:endoscopic_video}
\end{figure}

The procedure was completed successfully. The robot maintained endoscopic visualization throughout the sphenoidectomy, inserting and maneuvering the endoscope within the narrow surgical corridor while the surgeon achieved the procedural objective. To our knowledge, this represents the first demonstration of a humanoid robot providing functional assistance through a surgical procedure. Representative endoscopic views obtained during the cadaveric demonstration are shown in Fig~\ref{fig:endoscopic_video}.

Thematic analysis of the post-procedure interview identified both strengths and areas requiring development.

Regarding task performance, system stability and precision met or exceeded expectations for sustained visualization. The surgeon characterized overall performance as ``better than what I expected, but nowhere near the quality of when I see a human doing that procedure.'' The humanoid maintained stable endoscope positioning throughout tissue removal without drift or tremor artifacts. 

Force control during tissue contact, however, was inadequate. The surgeon noted endoscopic artifacts indicating excessive pressure: ``You can start to see artifacts in the endoscope when the robot was applying a significant amount of pressure onto the surrounding structures, and for a live patient, that would cause issues with risk.'' Human assistants possess proprioceptive awareness that ``when they are applying pressure to the patient, they're going to cause harm.''  Whether this limitation reflects the platform, the interface, or operator inexperience remains unclear; the teleoperation system lacks haptic feedback, and operator learning curve may contribute. Resolving this will require haptic feedback integration, autonomous force regulation, or systematic training studies.

Form factor and workspace presented distinct challenges. The surgeon noted: ``The endoscope is usually placed at the periphery of that corridor to give more dexterity and more degrees of freedom of the movement of the surgical instruments. The problem with the robot right now is that it's taking up the isocenter of that corridor.'' This constraint likely reflects initial design choices, including robot positioning, mount geometry, and approach angle, rather than fundamental platform limitations. However, some constraints may persist because humanoid kinematics cannot fully replicate the adaptive repositioning of a human assistant. Future iterations should explore alternative configurations to determine how much of this limitation is addressable.

Near-term clinical value is strongest for autonomous diagnostic scoping and scrub-tech roles. Automated endoscopic examination, particularly at the start of cases or in remote settings, offers high utility: ``They could perform this endoscopy or bronchoscopy anywhere in the world.'' A multi-camera assistant that manages and passes instruments is attractive and more tractable than operative autonomy: ``Replacing the scrub tech with a robot that can grab the instruments and hand them to you is particularly interesting.''

Trust and safety priorities are clear: eliminate erratic movements, provide pre-failure warnings, and implement a graceful abort with controlled withdrawal from the patient. As emphasized, ``An elegant way to abort the operation is the most important aspect,'' and ``it needs to communicate whenever there’s going to be a fault and it needs to exit... if it has to pull out of the patient, I can’t continue to operate.'' Autonomy for endoscope navigation was viewed as tractable and high-impact: ``I don’t think it’s a very hard problem to teach the robot to move itself... if it does it automatically, I think that’s a home run.''

Interaction and control findings revealed the importance of shared conventions. Establishing patient-relative orientation and a common language was essential: ``In medicine, we always give directions relative to the patient... we had to first figure out a common language.'' Practice sessions improved coordination substantially. For autonomous operation, the desired behavior is human-like situational awareness that minimizes verbal commands: ``the robot would constantly make the adjustments, and it would behave exactly like a human.''

Relative to large, rigid surgical platforms, a humanoid assistant is appealing because OR workflows are designed around multi-human teams, and a human-like assistant can integrate with minimal reconfiguration. The humanoid's potential advantage lies not in high precision. Rather, its value lies in versatility and compatibility with human-designed environments and workflows. Unlike other surgical robotic systems, which have ``a very large, cumbersome form factor'' requiring careful placement, the humanoid could theoretically ``integrate very easily into the surgical workforce, because I've already integrated another human.'' This suggests complementary rather than competing roles for humanoid and other surgical robotic approaches.

This proof of concept establishes that the humanoid form factor can meet the physical demands of surgical assistance. The teleoperated robot inserted and maneuvered the endoscope within a confined anatomical workspace, maintained visualization, and enabled procedure completion. The cognitive demands, including situational awareness, procedural anticipation, and responsive adjustment, were supplied by the teleoperator. The gap between this demonstration and clinical deployment remains substantial. Force feedback is absent, workspace geometry constrained instrument access, and coordination required extensive operator training. These challenges are tractable but non-trivial. Developing autonomous systems that replicate the capabilities of trained human assistants represents the key opportunity ahead. This work provides initial evidence that humanoid surgical assistance warrants continued investigation and characterizes specific engineering targets for that development.

\section{Methods}\label{method}
\subsection{Study Design}
This proof-of-concept study combined a procedural demonstration with qualitative evaluation to assess the feasibility of the humanoid form factor for endoscopic visualization in sinus surgery. An attending otolaryngologist with over 15 years of experience in endoscopic sinus surgery performed a sphenoidectomy on a cadaveric specimen, while a teleoperator controlled the humanoid-mounted endoscope to provide visualization.
We employed teleoperation to isolate embodiment and manipulability constraints, with questions of autonomous capability reserved for future work.

During the sphenoidectomy, the humanoid-mounted endoscope was required to navigate to the surgical region of interest, then maintain stable, correctly oriented visualization for unobstructed surgical tool access during tissue removal. Following tissue ablation, the endoscope was repositioned to visualize newly exposed anatomical regions and confirm adequate exposure.
Feasibility was defined as successful completion of key procedural steps with stable endoscopic visualization. Endoscopic video was recorded continuously for documentation and qualitative analysis. Video was also captured from an external vantage point to document the full procedural workflow.

The qualitative evaluation followed a semi-structured post-completion interview with the surgeon lasting approximately 30 minutes. The interview protocol compromised five domains: task performance; form factor and workspace, clinical applications, trust and safety, and interaction and control, where each domain included primary questions with standardized probes. The session was audio-recorded and transcribed for thematic analysis.

\subsection{Demonstration Setup}
The cadaveric sphenoidectomy was conducted in a research OR with the specimen positioned for endoscopic sinus surgery (Fig.~\ref{fig:overview}). The humanoid robot was positioned adjacent to the surgical table such that the surgeon maintained sufficient working space at the surgical table for standard instrument manipulation. A standard rigid endoscope (Karl Storz Image1 HD camera system) was mounted to the robot's right arm via a custom 3D-printed rigid adapter. The teleoperator maintained continuous control of the endoscope positioning through upper-body motion as no base repositioning was required during the procedure. The endoscopic video feed was visible to both the surgeon and the teleoperator, and verbal communication was maintained throughout to facilitate coordination.

\subsection{Robot System Description}

The robotic platform used in this study was the G1 humanoid robot (Unitree Robotics), which has 29 degrees of freedom (DoF). The G1 features a compact anthropomorphic form factor, with each upper limb providing 7-DoF manipulation capability. For this demonstration, the robot was secured to a chair to immobilize the lower body and trunk, and only upper-body motion was enabled. A standard surgical endoscope was mounted to the robot’s right end effector using a rigid adapter, which provided stable attachment while preserving wrist articulation for fine orientation control.

Teleoperation was implemented through a mixed reality interface using a Meta Quest 3 head-mounted display. The robot camera video stream was transmitted in real time to the headset and displayed as a virtual display within the operator’s passthrough view, enabling visualization of the surgical workspace while maintaining situational awareness. Motion commands for the G1’s upper limb were provided via handheld controllers, allowing continuous control of the endoscope pose (position and orientation). Specifically, the controller poses were mapped to the desired robot end effector poses, and joint-level commands were computed using inverse kinematics to generate the corresponding joint targets for the robot.

A trained researcher operated the system throughout the procedure. Prior to the experiment, the operator completed practice sessions for endoscopic guidance both manually and via the humanoid platform, establishing shared communication conventions with the supervising surgeon to facilitate intraoperative coordination.



\backmatter

\section{Supplementary Information}

\begin{figure}[H]
    \centering
    \includegraphics[width=0.8\linewidth]{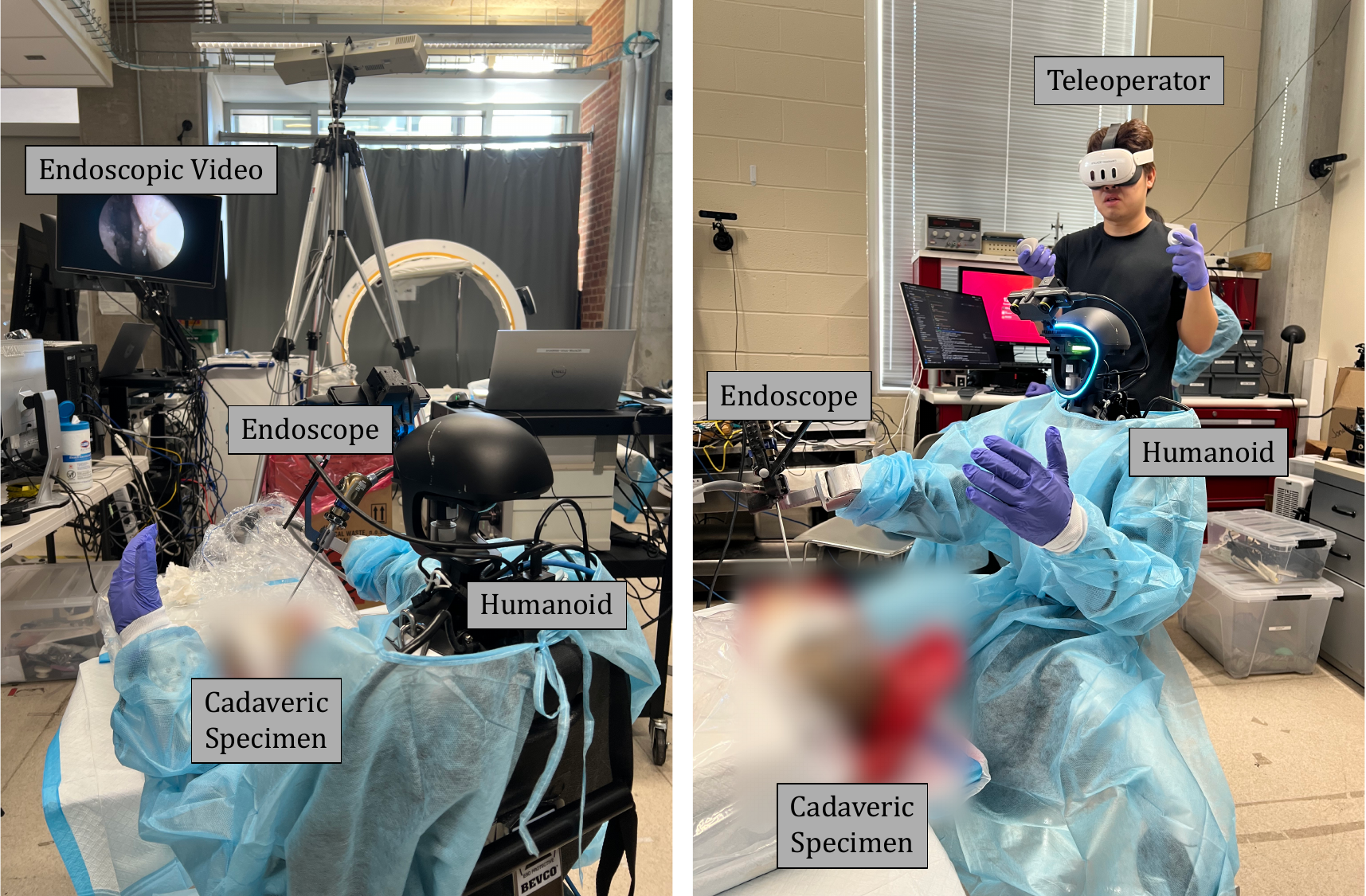}
    \caption{Cadaveric demonstration of the humanoid-assisted endoscopic setup. }
    \label{fig:cadaver}
\end{figure}







\section{Declarations}


\begin{itemize}
\item \textbf{Funding:} This work was funded in part by NIH R01EB030511 and NSF No.~2239077 in part by Johns Hopkins University internal funds. The content is solely the responsibility of the authors and does not necessarily represent the official views of the National Institutes of Health.

\item \textbf{Ethical approval} All data collection procedures in this study were reviewed and approved by the appropriate Institutional Review Board (IRB00267324) and were conducted in accordance with relevant ethical guidelines and regulations.

\end{itemize}

\bibliography{sn-bibliography}

\end{document}